# A Critical Note on the Evaluation of Clustering Algorithms

**Tiantian Zhang**
2573546543@qq.com
International Graduate School of
Tsinghua University

**Li Zhong**
zhongl18@mails.tsinghua.edu.cn
International Graduate School of
Tsinghua University

**Bo Yuan**
yuanb@sz.tsinghua.edu
International Graduate School of
Tsinghua University

## Abstract

Experimental evaluation is a major research methodology for investigating clustering algorithms and many other machine learning algorithms. For this purpose, a number of benchmark datasets have been widely used in the literature and their quality plays a key role on the value of the research work. However, in most of the existing studies, little attention has been paid to the properties of the datasets and they are often regarded as black-box problems. For example, it is common to use datasets intended for classification in clustering research and assume class labels as the ground truth for judging the quality of clustering. In our work, with the help of advanced visualization and dimension reduction techniques, we show that this practice may seriously compromise the research quality and produce misleading results. We suggest that the applicability of existing benchmark datasets should be carefully revisited and significant efforts need to be devoted to improving the current practice of experimental evaluation of clustering algorithms to ensure an essential match between algorithms and problems.

## Introduction

Clustering is one of the fundamental research areas in data science and has found numerous applications in a wide range of domains such as e-commerce, custom relationship management, image processing, and bioinformatics (Sarwar et al. 2002; Ngai, Xiu and Chau 2009; Dhanachandra, Manglem and Chanu 2015; Wiwie, Baumbach and Röttger 2015; Zou et al. 2018). In a general sense, a cluster refers to a group of data objects that are densely connected (Ester et al. 1996) and the task of clustering is to identify all such clusters from a given dataset. In practice, the challenge of clustering analysis mainly comes from factors such as irregular clusters, strong noise and high dimensionality (Zhang and Yuan 2018).

As an unsupervised learning paradigm, clustering does not require the original data to be labelled in advance, which may be very expensive in real-world scenarios. Instead, it aims at automatically exploring the inherent structure of the datasets to help people acquire an in-depth appreciation of the key properties of the data.

Similar to the field of classification, a common practice for investigating clustering techniques is by empirical studies where a set of benchmark datasets are used to quantitatively evaluate the performance of specific algorithms. As a result, it is clear that the quality of benchmark datasets plays a key role on the validness or effectiveness of the research outcomes.

To comprehend the current standard of clustering research, we reviewed a number of representative literatures, including survey papers (Jain, Murty and Flynn 1999; Xu and Wunsch 2005; Xu and Tian 2015) and some recent publications in leading journals and conferences (Maurus and Plant 2016; Zhu, Ting and Carman 2016; Zhu et al. 2017; Bojchevski and Günnemann 2018; Wang et al. 2019). There are mainly two categories of datasets in use: i) synthetic datasets, which are often of low dimensions for illustration purpose; ii) real-world datasets, which can be flexible in terms of size and dimensionality.

Normally, one or two synthetic 2D or 3D datasets are adopted to demonstrate the core procedure and mechanism of the proposed clustering algorithms, as they are relatively easy to be visualized. After that, real-world datasets come into play to provide further evidence on the practical performance of the algorithms, as it is often assumed that as long as real-world datasets are in use, it is reasonably plausible to make conclusive claims.

Note that in both cases "true" cluster labels are needed to judge the quality of clustering algorithms. For instance, for a 2D or 3D dataset, it is possible to visually identify its clustering pattern (e.g., the number of clusters and the membership of each data object). However, for higher dimensional datasets, it would be very challenging to intuitively make similar judgement due to difficulties in visualization. A key implication is that, for non-trivial datasets, it is rarely the case that true cluster labels are readily available. Consequently, people often take a "short-cut" by directly using classification datasets such as those in the UCI repository (Dua and Graff 2019) while assuming their existing class labels as the ground truth for clustering.

Unfortunately, as this paper will point out, it is a serious flaw in clustering research that has been prevalent for many years without any sign of decease. We claim that this defective research methodology has significantly compromised the quality of clustering research, resulting in inaccurate or misleading results. The key issue is that those labels are defined for classification purpose, not clustering, and mixing the two scenarios without any clear justification can produce unpredictable consequences.

For example, for classification purpose, a 2-class dataset can be created by collecting a set of features (e.g., height and weight) from male and female subjects, respectively, and assigning the label "male" or "female" to each corresponding data record. In this case, the label itself can be regarded as the ground truth against which to evaluate the outcomes of classification models.

However, this type of labeling is only an indicator of the class property instead of the distribution of the data, which is the main concern of clustering analysis. For example, as shown in Figure 1(a), data records with the same class label (circle) are not necessarily densely connected to each other. Instead, it is more reasonable to split them into two clusters. By contrast, in Figure 1(b), all data records are densely connected to each other and thus should form a single cluster, regardless of their class labels.

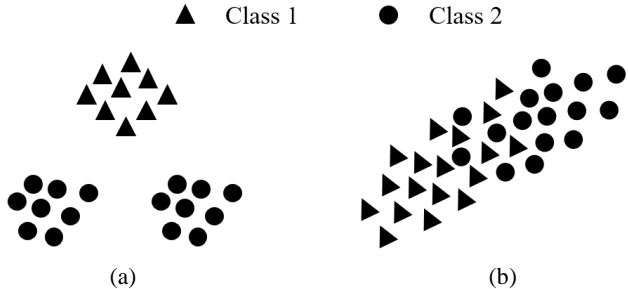

Figure 1: (a) a 2-class dataset where data objects with the same class label are distant from each other and should be split into different clusters; (b) a 2-class dataset where all data objects are densely connected to each other and thus form a single cluster.

The major motivation of our work is to improve the current practice of empirical research in clustering analysis and the key contributions of our paper are as follows:

- To highlight the importance of benchmark datasets and raise the alarm about the current practice of evaluating clustering algorithms;
- To show the potential issues of using class labels as the ground truth in the evaluation of clustering algorithms, using a synthetic dataset;
- To show that some popular classification datasets used in clustering research can produce misleading results and must be carefully revisited, using a set of in-depth case studies.

It should be noted that, unlike many other studies focusing on proposing new algorithms and performing competitive evaluation, the objective of our work is not to compare specific clustering algorithms. Furthermore, it is not our intention to make any general claim on the performance of the algorithms or the quality of the metrics involved. Instead, we are dedicated to conducting controlled experiments to reveal the hidden factors that are critical yet not well exposed to the research community. We are also fully aware of the unprecedented challenges associated with conducting comprehensive analysis on all major benchmark datasets, which will be pursued in the future.

In the next, we briefly introduce the metrics used in our work, followed by three case studies to show why it is not technically sound to use class labels as the ground truth for clustering as well as the possible consequences. This paper is concluded with some formal analysis and further discussions on the better practice of clustering research.

## Preliminaries

### Clustering

Clustering is the task of grouping a set of objects so that objects in the same group (called a cluster) are more similar (in some sense) to each other than to those in other groups (other clusters) (Wikipedia contributions 2019).

Given a set of $n$ elements (data objects) with $d$ features:

$$\mathcal{X} = \{(x_1, x_2, \ldots, x_n) | x_i \in R^d \text{ for } 1 \leq i \leq n\}$$

then, a clustering pattern is a partition of $\mathcal{X}$ into $k$ nonempty disjoint subsets $\mathcal{C} = \{\mathcal{C}_1, \mathcal{C}_2, \mathcal{C}_3, \ldots, \mathcal{C}_k\}$ such that:

1. $\forall \mathcal{C}_i, \mathcal{C}_j \in \mathcal{C}$, if $i \neq j$, then $\mathcal{C}_i \cap \mathcal{C}_j = \emptyset$;
2. $\cup_{i=1}^{k} \mathcal{C}_i = \mathcal{X}$.

Each clustering pattern specifies a sequence of cluster centroids and sizes. Let $c_i$ be the centroid and $n_i = |\mathcal{C}_i|$ be the size of the $i$th cluster, respectively. The sequence of cluster centroids is $[c_1, c_2, \ldots, c_k]$ and the sequence of cluster sizes is $[n_1, n_2, \ldots, n_k]$, subject to:

$$\sum_{i=1}^{k} n_i = n \quad (1)$$

Throughout this paper, we assume that the ground truth of clustering over $\mathcal{X}$ is $\mathcal{V} = \{\mathcal{V}_1, \mathcal{V}_2, \mathcal{V}_3, \ldots, \mathcal{V}_{k_\mathcal{V}}\}$ with $k_\mathcal{V}$ clusters of centroids $c_{\mathcal{V}_i}$ and sizes $n_{\mathcal{V}_i} = |\mathcal{V}_i|$. The result of a clustering algorithm over the same data $\mathcal{X}$ is $\mathcal{U} = \{\mathcal{U}_1, \mathcal{U}_2, \mathcal{U}_3, \ldots, \mathcal{U}_{k_\mathcal{U}}\}$ with $k_\mathcal{U}$ clusters of centroids $c_{\mathcal{U}_j}$ and sizes $n_{\mathcal{U}_j} = |\mathcal{U}_j|$.

### Performance Metrics

Evaluation (or "validation") of clustering results is as difficult as the clustering itself (Pfitzner, Leibbrandt and

Powers 2009). Generally, the performance metrics for clustering can be divided into two categories:

1. Internal Criteria: focusing on the natural relationships among clusters, such as the compactness of each cluster and the separation between clusters, which do not require true cluster labels.

2. External Criteria: focusing on the distribution differences between clustering results and the existing "ground truth", which require true cluster labels.

Some well-known internal evaluation measures are described as follows, usually based on the intuition that objects in the same cluster should be more similar to each other than to objects in different clusters.

**Davies-Bouldin Index (DBI).** DBI (Davies and Bouldin 1979) is evaluated based on the maximum ratio of intra-cluster compactness to inter-class dispersion:

$$DBI = \frac{1}{k_U} \sum_{i=1}^{k_U} max_{i \neq j} \left( \frac{\sigma_i + \sigma_j}{d(c_{U_i}, c_{U_j})} \right) \quad (2)$$

where $d(c_{U_i}, c_{U_j})$ is the distance between centroids $c_{U_i}$ and $c_{U_j}$ chosen as the representatives of clusters $U_i$ and $U_j$, and $\sigma_i$ is the average distance of all objects in cluster $U_i$ to its centroid $c_{U_i}$. Since a good clustering pattern produces clusters with low intra-cluster distance (high intra-cluster similarity) and high inter-cluster distance (low inter-cluster similarity), the minimum DBI value (ranged from $[0, +\infty)$) indicates the natural partitions of datasets.

**Silhouette Coefficient (SC).** SC (Rousseeuw 1987) contrasts the average distance to objects in the same cluster with the average distance to objects in other clusters. Each cluster is represented by a silhouette and the entire clustering result is presented by combining the silhouettes into a single plot. The average silhouette width provides an estimation of clustering validity:

$$SC = \frac{1}{n} \sum_{i=1}^{n} SC(i) = \frac{1}{n} \sum_{i=1}^{n} \frac{b(x_i) - a(x_i)}{max\{a(x_i), b(x_i)\}} \quad (3)$$

where $a(x_i)$ is the average dissimilarity of $x_i$ to all other objects in the same cluster and $b(x_i)$ is the minimum of the average dissimilarities of $x_i$ to all objects in other clusters. The maximum SC value (ranged from $[-1, 1]$) indicates the most reasonable clustering of datasets.

External evaluation criteria are functions that measure the similarity of two assignments $U$ and $V$.

**Adjusted Rand Index (ARI).** ARI (Hubert and Arabie 1985) measures the difference between two clustering results, adjusted for the chance grouping of elements for Rand Index (RI) (Rand 1971):

$$ARI = \frac{RI - E[RI]}{max(RI) - E[RI]}, \quad RI = \frac{a+b}{\binom{n}{2}} \quad (4)$$

where $a$ is the number of paired objects placed in the same cluster in both partitions and $b$ is the number of paired objects placed in different clusters in both partitions. Intuitively, $a + b$ can be considered as the number of agreements between $U$ and $V$. $\binom{n}{2}$ is the number of object pairs in the dataset. The maximum ARI value (ranged from $[-1, 1]$) indicates the largest goodness of fit between the clustering result and the desired partition of data.

**Information Theoretic Based Measures.** The Mutual Information (MI) (Shannon 1948; Cover and Thomas 1991) is a symmetric measure that quantifies the mutual dependence between two random variables, or the information that two random variables share. In data mining, it can be used to determine the similarity of two clustering patterns $U$ and $V$ of a dataset $X$:

$$MI(U, V) = \sum_{i=1}^{k_U} \sum_{j=1}^{k_V} P(i,j) log \frac{P(i,j)}{P(i)P'(j)} \quad (5)$$

where $P(i,j)$ denotes the probability that a point belongs to both $U_i$ and $V_j$; $P(i)$ is the probability that the random object in $X$ falls into $U_i$; $P'(j)$ is the probability that the random object in $X$ falls into $V_j$.

There are two normalized versions of MI: Normalized Mutual Information (NMI) (Strehl and Ghosh 2002) and Adjusted Mutual Information (AMI) (Vinh, Epps and Bailey 2009), which is normalized against chance:

$$NMI(U, V) = \frac{MI(U,V)}{\sqrt{H(U)H(V)}} \quad (6)$$

$$AMI(U, V) = \frac{MI(U,V) - E\{MI(U,V)\}}{max\{H(U), H(V)\} - E\{MI(U,V)\}} \quad (7)$$

where $MI(U, V)$ is the MI between two partitions and $H(U)$ and $H(V)$ are the entropy values. The maximum MI/NMI value (ranged from [0, 1]) and AMI (ranged from [-1, 1]) value both indicate the most significant agreement between two clustering patterns.

## Synthetic Dataset: A Step Forward

In this section, we used a 2D synthetic 2-class dataset to demonstrate that it is not technically sound to use the class labels of classification datasets as the ground truth for evaluating clustering algorithms.

As shown in Figure 2(a), the dataset *SD_2* was created within $[0, 1]^2$. Class 1 (triangles) contained around 800 objects following a Gaussian distribution with standard deviations $[0.008, 0.002]$. Class 2 (circles) contained two Gaussian distributions with around 1000 (left) and 1200 (right) objects with standard deviations $[0.007, 0.003]$ and $[0.007, 0.004]$, respectively.

We used K-means (Hartigan and Wong 1979) and DBSCAN (Ester et al. 1996) as two representative clustering algorithms in the experiment. Intuitively, we can see that the dataset contained three clusters. Therefore, we set $k = 3$ for K-means, and the result is shown in Figure 2(b, right). Meanwhile, for DBSCAN, the commonly adopted method *k-dist graph* was used to determine its key parame-ters: $Eps = 0.025$, $minPts = 4$, and the result is shown in Figure 2(c, right). Table 1 (fourth and sixth rows) summarizes the values of the corresponding performance metrics, including internal criteria (DBI and SC) and external criteria (ARI, MI, NMI and AMI), using the class labels as the ground truth of the clustering directly.

In some previous studies (Maurus and Plant 2016), in order to achieve the maximum agreement between clustering results and the ground truth, researchers usually traverse the parameters within a certain range and take the best clustering result as the final result. Following this practice, we systematically varied $k$ of K-means with $k = \{1, 2, 3\}$ and $Eps$ of DBSCAN (fixed $minPts = 4$) with $Eps = \{0.010, 0.015, …, 0.040\}$, and presented the best AMI results in the third and fifth rows of Table 1. The corresponding clustering results are visualized in the left figures of Figure 2(b) and Figure 2(c), respectively.

According to the data distribution of $SD\_2$ as shown in Figure 2(a), it is clear that: i) for K-means, the clustering result with $k = 3$ is more reasonable than that with $k = 2$; ii) for DBSCAN, given $minPts = 4$, the clustering result with $Eps = 0.025$ is more reasonable than that with $Eps = 0.035$.

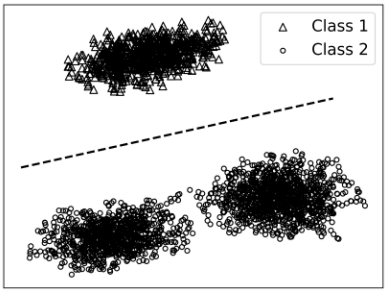

(a) Raw data: binary classification

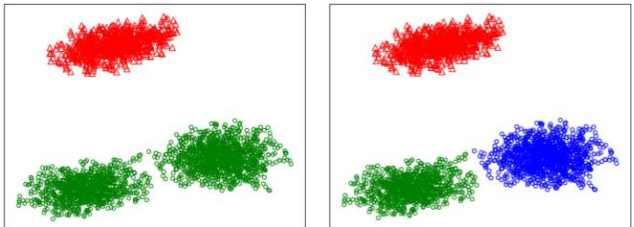

(b) K-means: $k = 2$ (left), $k = 3$ (right)

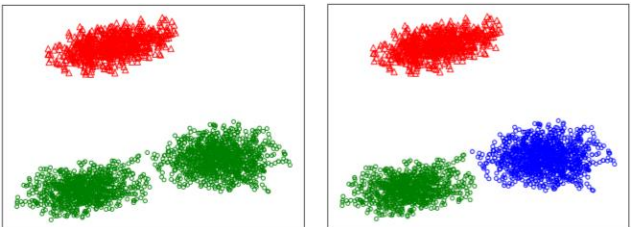

(c) DBSCAN: $minPts = 4$, $Eps = 0.035$ (left), $Eps = 0.025$ (right)

Figure 2: The synthetic dataset $SD\_2$. (a) raw data; (b) the clustering results of K-means; (c) the clustering results of DBSCAN.

However, significant inconsistency can be readily observed from Table 1. For example, in terms of DBI and SC, their values can support our above claim (the smaller the DBI value and the larger the SC value, the better the clustering quality). However, in terms of the external criteria such as ARI, NMI, and AMI that depend on the ground truth, their values tell a different story. For example, the clustering result with $k = 2$ for K-means was better than that with $k = 3$, even when the original dataset clearly contained three clusters instead of two clusters. Similar conclusion can be also observed from the result of DBSCAN.

Note that the above contradiction is not due to the specific clustering algorithms or the applicability of the performance criteria. Instead, it shows that, on such a simple dataset, incorrect clustering results can be obtained due to a "chain of errors": i) class labels are not necessarily consistent with the clustering pattern of a dataset; ii) class labels are used as the ground truth for performance metrics; iii) clustering algorithms that produce high scores in terms of these metrics are regarded as being superior to others, even if they produce completely misleading results.

Table 1: The metric scores for clustering on $SD\_2$

|  | DBI | SC | ARI | MI | NMI | AMI |
|---|---|---|---|---|---|---|
| Class Label | 0.554 | 0.608 | - | - | - | - |
| K-means (k=2) | 0.554 | 0.608 | **1.000** | 0.579 | **1.000** | **1.000** |
| K-means (k=3) | **0.347** | **0.746** | 0.499 | 0.579 | 0.695 | 0.695 |
| DBSCAN (Eps=0.035) | 0.554 | 0.608 | **1.000** | 0.579 | **1.000** | **1.000** |
| DBSCAN (Eps=0.025) | **0.347** | **0.746** | 0.499 | 0.579 | 0.695 | 0.695 |

## Real-World Datasets: Overlapping Data

In this section, we focus on the real-world situations where groups of data with different class labels may overlap with each other. From the perspective of clustering, it means that it may be more reasonable for some groups to be merged into a single cluster. Consequently, using class labels as the ground truth for clustering evaluation is not appropriate in this situation.

*Weight-height*[1] is a 2D dataset containing 10,000 records of the weights and heights of male and female subjects. For clarity, we used 400 samples randomly sampled from the raw data in the following experiments, as shown in Figure 3(a) where triangles represent the class *male* and circles represent the class *female*.

It is clear that there is some overlapping region between the two classes, making the distribution of the data more like a single Gaussian distribution than two separate Gaussian distributions. Based on this observation, we set $k = 1$ to perform K-means clustering, and the clustering result is shown in Figure 3(b, left). Meanwhile, for DBSCAN, *k-dist graph* was again used to determine its key parameters: $Eps = 0.070$, $minPts = 4$, and the clustering result is shown in Figure 3(c), where all objects were grouped into a single cluster with two noise data points. However, as shown in Table 2 (third and fifth rows), since these two clustering results are both significantly inconsistent with the class labels of the original data, their external metric values are all close to zero.

---

[1] The dataset *weight-height* is from Kaggle, and the download link is https://www.kaggle.com/mustafaali96/weight-height.

To elucidate the point further, we performed K-means clustering with $k = 2$ in consistence with the class labels of the raw data. Figure 3(b, right) shows the corresponding clustering result where all objects in the dataset were grouped into two non-overlapping clusters. In this case, the metric scores have been significantly improved compared with the situation when $k = 1$, as shown in the fourth row of Table 2. Note that even when the dataset was divided into two clusters with $k = 2$, since each cluster contained objects with different class labels, the external metric values were still only moderate, underestimating the true performance of the clustering algorithms.

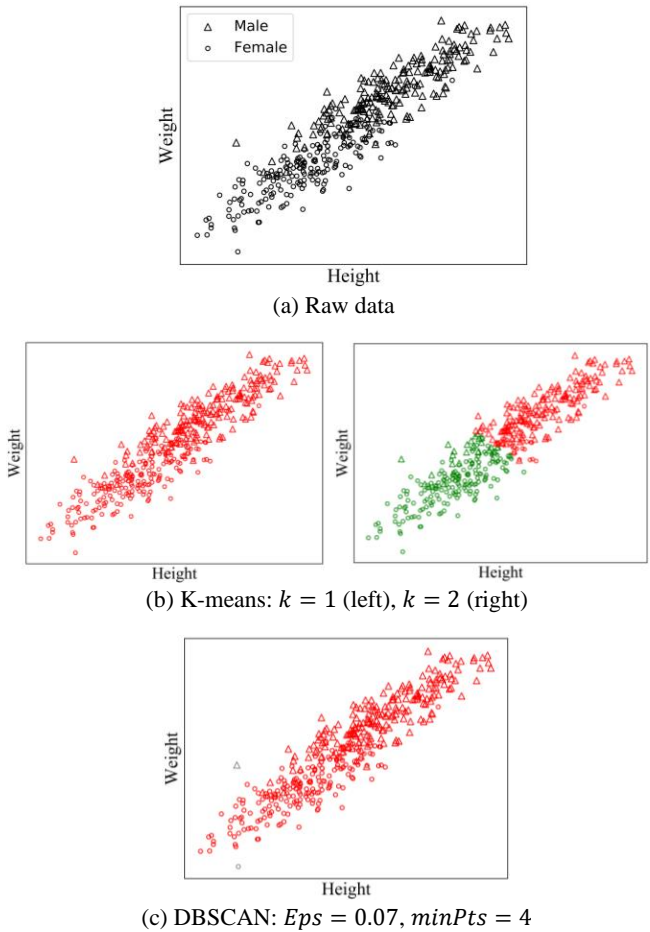

Figure 3: The *Weight-Height* dataset. (a) raw data; (b) the clustering results of K-means; (c) the clustering result of DBSCAN.

In summary, in a dataset where objects with different class labels significantly overlap with each other, multiple classes may produce a single cluster and use the class labels to measure the quality of clustering is clearly unjustified.

Table 2: The metric scores for clustering on *Weight-Height*

|  | DBI | SC | ARI | MI | NMI | AMI |
|---|---|---|---|---|---|---|
| Class Label | 0.759 | 0.430 | - | - | - | - |
| K-means (k=1) | -[2] | - | 0.000 | 0.000 | 0.000 | 0.000 |
| K-means (k=2) | 0.632 | 0.531 | 0.569 | 0.322 | 0.464 | 0.463 |
| DBSCAN (Eps=0.070) | 1.050 | 0.258 | 0.000 | 0.000 | 0.000 | -0.005 |

### Real-World Datasets: Splitting Data

Another possibility in the real word is that objects with the same class label may correspond to more than one densely connected regions separated by sparse-density regions. As far as clustering is concerned, it is reasonable to regard these separate dense regions as independent clusters, even if they share the same class label. Therefore, it is also not technically sound to evaluate the performance of clustering algorithms using the class labels as the ground truth.

*Accelerometer*[3] is a publicly collected dataset for Activities of Daily Living (ADL) recognition and classification, which is composed of the recordings of 16 volunteers performing 14 simple activities (e.g., brush_teeth, climb_stairs, descend_stairs, drink_glass, etc.) while carrying a single wrist-worn tri-axial accelerometer. Each record consists of three attributes, which represent the acceleration along the x axis, y axis and z axis of the accelerometer when the volunteers were performing their daily activities. For clarity, we selected the data of two activities (270 records labeled *climb_stairs* and 594 records labeled *descend_stairs*) from the original dataset for the following analysis.

In Figure 4(a), triangles represent data objects labeled *climb_stairs*, and circles represent data objects labeled *descend_stairs*. It is clear that the class *descend_stairs* is split into roughly two densely connected regions and one of them overlaps with the class *climb_stairs*.

Intuitively, from the perspective of clustering, part of the data objects labeled *descend_stairs* located in the upper left corner should be considered as an independent cluster, while the rest data objects of the same class located in the lower right corner that overlap with the class *climb_stairs* should be merged together as a single cluster. To further verify this observation, we run K-means with $k = 2$ and DBSCAN with $Eps = 0.110$, $minPts = 5$. As a result, K-means produced two clusters with one cluster containing objects from both classes, as shown in Figure 4(b). Meanwhile, in Figure 4(c), DBSCAN produced similar results.

---

[2] As internal criteria, DBI and SC only make sense when there are at least two clusters.
[3] The dataset *Accelerometer* is from UCI Machine Learning Repository: https://archive.ics.uci.edu/ml/datasets/Dataset+for+ADL+Recognition+with+Wrist-worn+Accelerometer.

In addition, the corresponding scores of various metrics are presented in Table 3. In terms of internal criteria (DBI and SC), the results of K-means and DBSCAN were much better than directly using class labels as the cluster labels. However, in terms of external criteria (ARI, MI, NMI and AMI), since we used class labels as the ground truth of clustering, both K-means and DBSCAN produced very inferior results, which were not consistent with the intuitively good clustering results in Figure 4.

This case study highlights again the issue of using class labels as the ground truth for clustering research when data objects with the same class label are split into multiple separate regions. In this situation, good metric scores do not necessarily correspond to natural clustering patterns

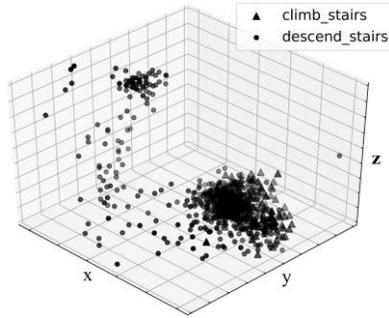

and may be misleading for algorithm evaluation.

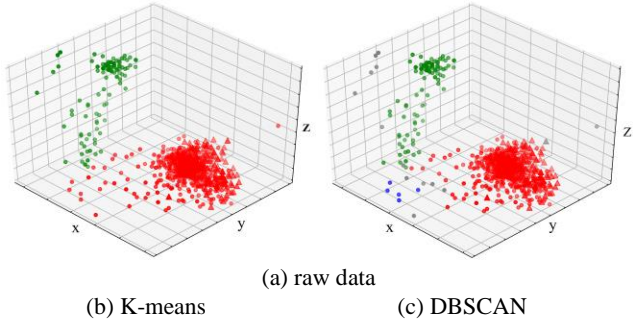

(a) raw data
(b) K-means          (c) DBSCAN

Figure 4: The *Accelerometer* dataset. (a) raw data; (b) the clustering result of K-means ($k = 2$); (c) the clustering result of DBSCAN ($Eps = 0.110$, $minPts = 5$).

Table 3: The metric scores for clustering on *Accelerometer*

|  | DBI | SC | ARI | MI | NMI | AMI |
|---|---|---|---|---|---|---|
| Class Label | 2.176 | -0.035 | - | - | - | - |
| K-means (k=2) | 0.492 | 0.702 | -0.070 | 0.055 | 0.109 | 0.108 |
| DBSCAN (Eps=0.110) | 1.533 | 0.600 | -0.078 | 0.058 | 0.104 | 0.101 |

For higher dimensional classification datasets, objects with the same class label are more likely to correspond to multiple clusters due to the sparsity of data in high-dimensional spaces.

*Vertebral Column*[4] is a biomedical dataset containing 310 instances with 6 biomechanical features used to classify orthopedic patients into three classes (*normal*, *disk hernia* and *spondylolisthesis*) or two classes (*normal* and *abnormal*). In this paper, we focused on the latter case, which merged the classes of *disk hernia* and *spondylolisthesis* into a single class *abnormal*.

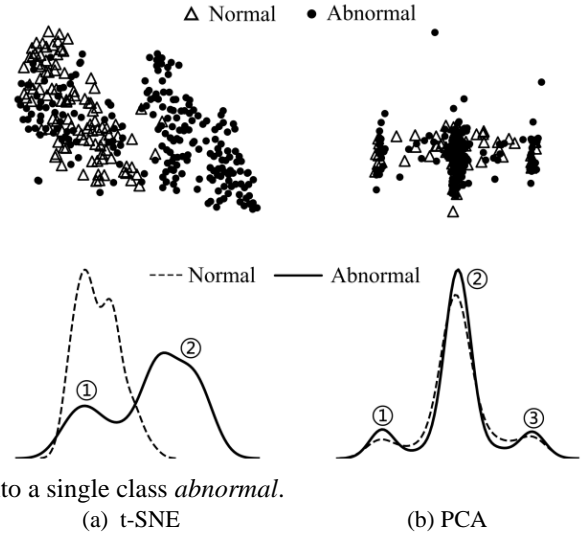

(a) t-SNE          (b) PCA

Figure 5: The visualization of *Vertebral Column* (6D, 2 classes). (a) the 2D plot and 1D histograms using t-SNE; (b) the 2D plot along the second and sixth principle components and 1D histograms using PCA.

Figure 5 shows the visualization of *Vertebral Column* using dimension reduction techniques t-SNE[5] (Maaten and Hinton 2008) and PCA[6] (Pearson 1901). The upper figure in Figure 5(a) shows the visualization of the data reduced to a 2D space using t-SNE where triangles represent *normal* and circles represent *abnormal*. The lower figure in Figure 5(a) presents the density curves of the two classes along the horizontal axis.

The upper figure in Figure 5(b) shows the projected 2D data using PCA where the horizontal axis is the sixth principal component and the vertical axis is the second principal component. These two dimensions were selected purposefully to show that objects belonging to the same class may be split into multiple regions. The lower figure in Fig-ure 5(b) presents the density curves of the two classes along the horizontal axis (the sixth principal component).

---

[4] The dataset *Vertebral Column* is from UCI Machine Learning Repository: https://archive.ics.uci.edu/ml/datasets/Vertebral+Column#.
[5] t-SNE (t-Distributed Stochastic Neighbor Embedding) is a nonlinear dimension reduction and visualization technique for exploring high-dimensional data, which finds the hidden patterns in data through the clusters identified based on the similarity between data objects in high-dimensional space, and can well retain the local and global structure of high-dimensional data.
[6] PCA (Principal Component Analysis) is one of the most widely used linear dimension reduction technique for retaining the most importance features of high-dimensional data.

From Figure 5(a), we can see that data objects in class *abnormal* was split into two separate dense regions in the 2D subspace after dimension reduction by t-SNE and the density curve also shows two distinct density peaks corresponding to these two dense regions. This means that, in the original space, it is sure that objects in class *abnormal* are not distributed in the form of a single cluster. The results of PCA provided more informative details, subdividing class *abnormal* into one dense region containing most of the data objects and two separate smaller dense regions, as shown in Figure 5(b). In this situation, according to the analysis at the beginning of this section, it is not justified to apply the class labels of this dataset to evaluating clustering performance.

## Discussion

In the above case studies, it is clear that when two subsets of data with different class labels overlap with each other, they can no longer be regarded as individual clusters and it is not appropriate to use the class information as the ground truth for clustering. Assume that there are $k$ classes in the dataset and each class of data is densely connected. An interesting question would be: if the location of each class is randomly distributed in the space, what is the probability that no class overlaps with any other class? If so, each class of data corresponds to a unique cluster and its class label can be used to indicate the cluster membership. In the next, we give a quantitative analysis in a 2D box-bounded space with width $w$ where each class is represented by a circle with identical radius $r$.

To make sure that two classes do not overlap with each other, the centers of the two circles must be more than $2r$ away from each other. If we randomly draw a pair of centers from the 2D space, the probability that the two circles do not overlap is:

$$p' \approx 1 - \frac{4r^2\pi}{w^2} \qquad (8)$$

Equation (8) is approximate as it does not give special consideration to circles that are close to the boundary of the space. In a space containing $k$ circles ($k \geq 2$), there are totally $\binom{k}{2}$ pairs of circles. By assuming that these events are independent, the probability that all circles are disjoint from others is:

$$p = (p')^{\binom{k}{2}} = \left(1 - \frac{4r^2\pi}{w^2}\right)^{\frac{k(k-1)}{2}} \qquad (9)$$

According to the Taylor series expansion of the exponential function, for $|x| \ll 1$, its first-order approximation is given by:

$$e^x \approx 1 + x \qquad (10)$$

So, for $r \ll w$, Equation (9) can be approximated as:

$$p \approx \left(e^{-\frac{4r^2\pi}{w^2}}\right)^{\frac{k(k-1)}{2}} = e^{-\frac{2r^2 k(k-1)\pi}{w^2}} \qquad (11)$$

Equation (11) indicates that, as the number of classes or the relative size of each class of data increases, the chance that some classes overlap with each other also increases accordingly. A set of simulations was conducted with $r = 0.01$, $w = 1.0$ and $k$ was systematically varied from 2 to 100. For each $k$ value, 10,000 trials were conducted and the non-overlapping probabilities are presented in Figure 6, which shows that it is more likely that some circles overlap with each other as $k$ increases, as indicated by Equation (11).

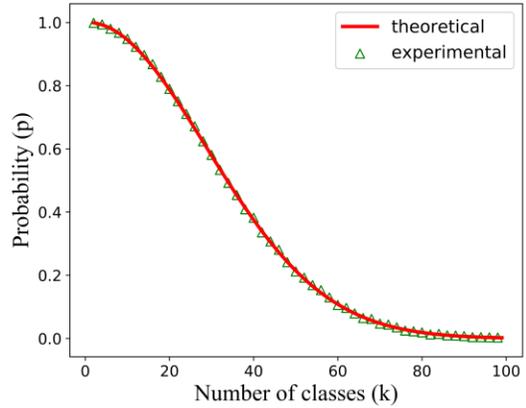

Figure 6: The probability ($p$) curve that all classes are disjoint from each other with regard to the number of classes ($k$).

## Conclusion

This paper calls for the close attention from the clustering research community on the current practice of empirical studies. In particular, we show that it is problematic to use classification datasets in clustering research without any *a prior* justification. We point out that class information is related to the property of each individual object while clustering is concerned about the relationship among objects and this inconsistence may result in serious issue. As shown in the case studies on both synthetic and real-world datasets, using class labels as the ground truth for clustering may produce contradicting and misleading results.

In practice, a single class may correspond to multiple clusters while multiple classes may be merged into a single class. Consequently, instead of arbitrarily choosing blackbox datasets from public repositories, it is highly recommended to have a clear understanding of the structure of datasets to provide at least the basic level of assurance about their applicability. Furthermore, due to the challenge of determining the true cluster labels for non-trivial real-world datasets, advanced synthetic datasets with controlled structure may need to be purposefully generated to better support the principled evaluation of clustering algorithms.